\documentclass[10pt,twocolumn,letterpaper]{article}

\usepackage{iccv}
\usepackage{times}
\usepackage{epsfig}
\usepackage{graphicx}
\usepackage{amsmath}
\usepackage{amssymb}
\usepackage{subfigure}
\usepackage{enumitem}

\usepackage[pagebackref=true,breaklinks=true,letterpaper=true,colorlinks,bookmarks=false]{hyperref}
\usepackage{tabularx}

\graphicspath{{./figures/}}
\newcommand{\rulesep}{\unskip\ \vrule\ }
\newcommand{\tabincell}[2]{\begin{tabular}{@{}#1@{}}#2\end{tabular}}
\newcommand{\norm}[1]{\left\lVert#1\right\rVert}

\iccvfinalcopy 
\def\iccvPaperID{2915} 

\ificcvfinal\fi

\begin{document}

\title{Disentangling Pose from Appearance in Monochrome Hand Images}

\author{Yikang Li$^{1}$\thanks{Work is done during an internship at Facebook Reality Labs.}, Chris Twigg$^{2}$, Yuting Ye$^{2}$, Lingling Tao$^{2}$, Xiaogang Wang$^{1}$\\
	{\tt\small \{ykli, xgwang\}@ee.cuhk.edu.hk \ \ \{chris.twigg, yuting.ye, lingling.tao\}@oculus.com} \\
	$^{1}$The Chinese University of Hong Kong, Hong Kong SAR, China\\ 
	$^{2}$Facebook Reality Labs,  USA 
}
\maketitle

\begin{abstract}
Hand pose estimation from monocular 2D image is challenging due to the
variation in lighting, appearance, and background.  While some success has been
achieved using deep neural networks, they typically require collecting a large dataset
that adequately samples all the axes of variation of hand images. It would therefore be useful
to find a representation of hand pose which is independent of the image  	
appearance~(like hand texture, lighting, background), so that we can 
synthesize unseen images by mixing pose-appearance combinations.
In this paper, we present a novel technique that disentangles the representation of pose from
a complementary appearance factor in 2D monochrome images.  We supervise this
disentanglement process using a network that learns to generate
images of hand using specified pose+appearance features.  Unlike previous work, we do not
require image pairs with a matching pose; instead, we use the pose annotations
already available and introduce a novel use of cycle consistency to ensure
orthogonality between the factors.  Experimental results show that our
self-disentanglement scheme successfully decomposes the hand image into pose
and its complementary appearance features of comparable quality as the method using paired data.
Additionally, training the model with extra synthesized images with unseen hand-appearance combinations by re-mixing pose and appearance factors from different images can improve the 2D pose estimation performance.
\end{abstract}

\begin{figure}[t!]
	\centering
	\subfigure[Variation in Hand Pose]{\label{fig:b}\includegraphics[width=0.23\textwidth]{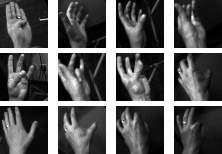}}
	\rulesep
	\subfigure[Variation in Appearance]{\label{fig:a}\includegraphics[width=0.23\textwidth]{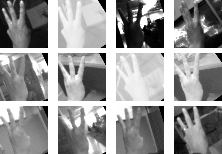}}
	\subfigure[Generated Images with Specified Pose and Appearance]{\label{fig:c}\includegraphics[width=0.47\textwidth]{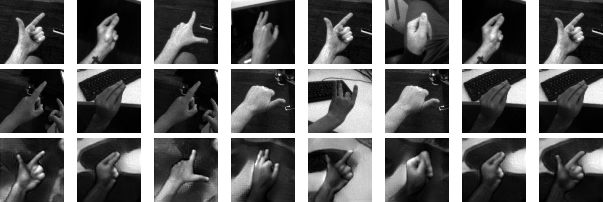}}
	\caption{Robust hand pose detection requires handling the large variation in (a) hand pose, and, (b) Image appearance, \eg different backgrounds, lighting conditions, camera exposures, hand textures, \etc. (c) Image generated with pose from the first row and appearance from the second row.}	\vspace{-10pt}
	\label{fig:variance}
	
\end{figure}

\section{Introduction}
Hand pose estimation is an important topic in computer vision
with many practical applications including virtual/augmented
reality~(AR/VR)~\cite{lee2009multithreaded,piumsomboon2013user} and 
human-computer
interaction~\cite{sridhar2015investigating,markussen2014vulture}. A large body
of work has shown robust hand pose estimation using RGB-D
cameras~\cite{hamer2009tracking,sridhar2016real,keskin2013real,tagliasacchi2015} or stereo
cameras~\cite{wang20116d,sridhar2013interactive} that provides 3D information about hands.  With the recent advance of deep learning techniques~\cite{imagenet_hinton,li2017scene,li2017vip,li2018visual,li2018factorizable}, researchers have begun exploring the use
of monocular 2D cameras ~\cite{zimmermann2017learning,mueller2018ganerated}, which are cheap and
ubiquitous thanks to their use in consumer devices like smart-phones and laptops. 

Despite recent success of applying deep learning in hand pose estimation from
monocular 2D images, there is still a substantial quality gap when comparing with depth-based approaches.  We believe that the culprit is the variability in hand
appearance caused by differences in lighting, backgrounds, and skin tones or
textures. The same hand pose can appear quite differently in daylight than
fluorescent lighting, and both harsh shadows and cluttered backgrounds tend to
confuse neural networks.  To ensure the robustness of neural networks, large amount of training data is typically required in order to adequately samples all the axes of variation.

In this work, we aim to improve the robustness of hand pose estimation from
monocular 2D images by finding a representation of hand pose that is 
independent of its appearance.
We propose to train a neural network that learns to ``disentangle'' a hand image into two
sets of features: the first captures the hand pose, while the second captures
the hand's appearance.  Pose features refer to the informative factors used to
reconstruct the actual hand pose (e.g.  the locations of the hand joints),
while the appearance feature denote the complementary ``inessential'' factors
of the image, such as the background, lighting conditions, hand textures, \etc.
We refer to this decomposition as Factor Disentanglement. 


Existing approaches to factor disentanglement generally require pairs of
annotated data~\cite{mathieu2016disentangling,gonzalez2018image}, where the
pairs share some features~(e.g. object class) but vary in others~(e.g.
background). These pairs supervise the disentanglement process, demonstrating
how different parts of the feature vector contribute to the composition of the
image.  While it is relatively easy to find multiple images that share the same
object class, finding pairs of images with identical hand but different appearance is considerably more challenging.  Instead, we propose to learn to disentangle the images using the supervision we do have: labeled poses for each training image.  

\begin{figure*}[t]
	\centering
	\includegraphics[width=0.95\textwidth]{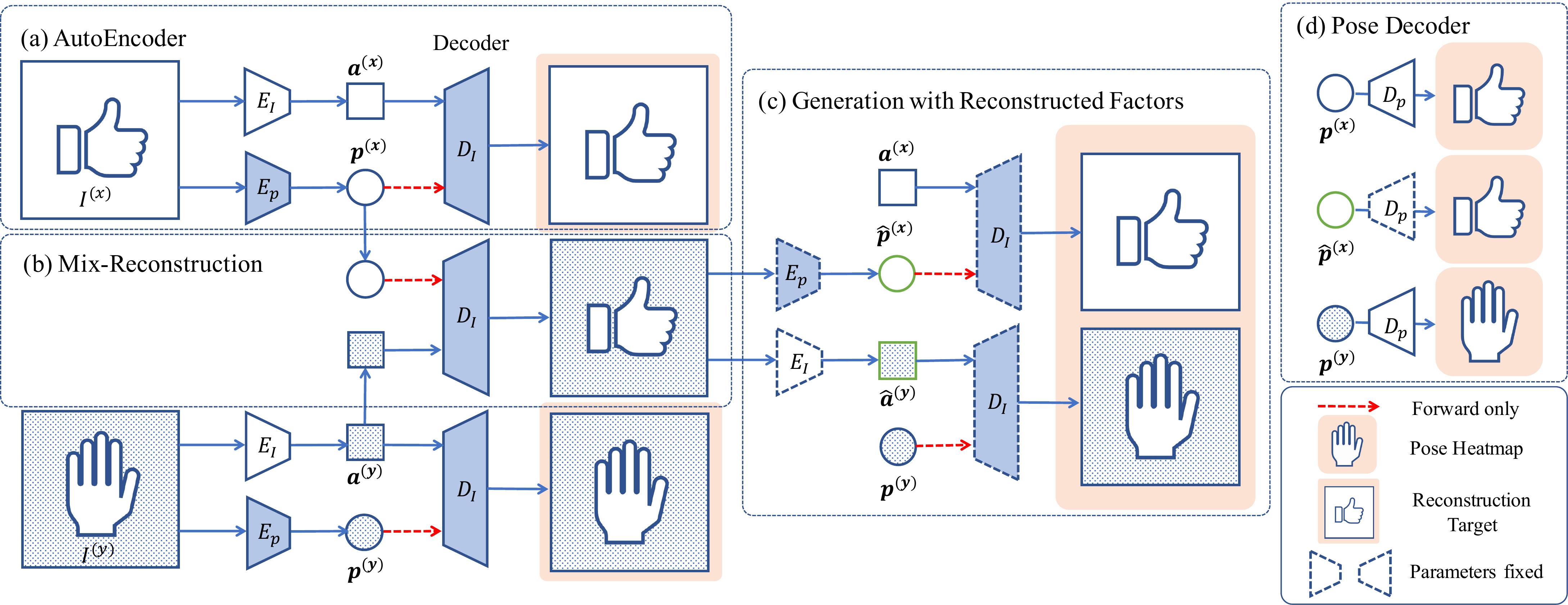}
	\caption{The overview of the self-disentanglement training process. 
		(a) Input images $I^{(x)}$ and $I^{(y)}$ are encoded into Pose and Appearance factors, 
		which contains the hand joint locations and its complementary image appearance information~(\eg background, lighting, hand texture, \etc) respectively. 
		$E_{p}$ and $E_{I}$ are encoders for the pose and appearance factors respectively. 
		Image decoder~$D_{I}$ is used to reconstruct the original images using the pose and appearance factors. 
		(b) We combine the pose factor from $I^{(x)}$ and appearance factor from $I^{(y)}$ to construct a ``mix-reconstructed'' hand image with expected pose and appearance. 
		(c) The mix-reconstructed image is decomposed back to the pose and appearance factors, 
		and the resulting pose (appearance) feature is combined with the original appearance (pose) feature to generate a new decoded image, 
		which should be similar to the original image. 
		(d) The pose factors are trained to predict the pose heatmap with the pose decoder $D_{p}$. 
		The dashed arrow indicates that we don't allow the gradients from the image reconstruction loss to back-propagate through the pose factors.
		The dashed-outlined modules mean they just work as an estimator to provide the gradients to the early stage. 
	}
	\label{fig:overview}
	\vspace{-3mm}
\end{figure*}

To do this, we start with the following principles: 
\begin{enumerate}
\item We should be able to predict the (labeled) hand pose using only the pose feature. 
\item We should be able to reconstruct the original image using the combination of pose plus appearance features.  
\end{enumerate}
However, this is not sufficient, because there is nothing that ensures that the
pose and appearance features are {\em orthogonal}, that is, they represent
different factors of the image.  The network can simply make the ``appearance''
feature effectively encode both pose and appearance, and reconstruct the image
while ignoring the separate pose component.  Therefore:
\begin{enumerate}[resume]
\item We should be able to combine the pose feature from one image with the
appearance feature from another to get a novel image whose pose matches the
first image and whose appearance matches the second. 
\end{enumerate}
Because we have no way to know {\em a priori} what this novel image should look
like, we can not supervise it with an image reconstruction loss.  Instead, we
use {\em cycle consistency}~\cite{zhu2017unpaired}: if we disentangle this
novel image, it should decompose back into the original pose and appearance
features. This will ensure that the network does not learn to encode the pose
into the appearance feature.  
We apply these three principles during our training process shown in Fig.~\ref{fig:overview}.

The proposed self-disentanglement framework is applied on a dataset of monochrome hand images. We show that learning to disentangle
hand pose and appearance features greatly improves the performance of hand pose estimation module in two ways: 1. the pose estimation module can learn a better pose feature representation when the factor disentanglement is learned jointly as an auxiliary task. 2. the dataset can be augmented by generating new images with different pose-appearance combinations during the process. Both methods lead to improvement over baseline on our hand pose dataset. In addition, we also show comparable results to a factor disentanglement network trained with the supervision of paired images. Due to the challenge of capturing perfect paired data, we resort to a synthetic dataset for this comparison, where a pair of hand images are rendered using path tracing~\cite{purcell2005ray} with identical hand pose but different background, lighting, and hand textures. Although our experiments are done using monochrome images, this framework can be easily extended for the case of RGB images.



The main contribution of this paper is as follows:
\begin{itemize}
  \item A self-distanglement network to disentangle the hand pose
  features from its complementary appearance features in monochrome images,
  without paired training data of identical poses.
  \item The proposed framework improves the robustness of supervised pose estimation to
  appearance variations, without use of additional data or labels.
\end{itemize}
 
\section{Related Work}

\textbf{Hand Tracking:}
Due to the large pose space, occlusions, and appearance variation such as lighting and skin tone, hand tracking from images is a challenging problem.  Methods that use multiple views~\cite{wang20116d,oikonomidis2011full,ballan2012motion} can deal with occlusions but are more difficult to set up.  A large body of work~\cite{hamer2009tracking,sridhar2016real,mueller2017real,tagliasacchi2015,tagliasacchi2015} has demonstrated high-quality tracking use depth/RGBD cameras.  While powerful, these sensors are still not largely available in consumer products.  More recently, there has been work on tracking hands from monocular RGB camera~\cite{zimmermann2017learning,mueller2018ganerated} using deep learning techniques.  In this work, we will focus on monochrome cameras due to their increased sensitivity to low light, but the methods should generalize to RGB camera case as well.

\textbf{Encoder-Decoder Structure and Autoencoder:} Our base architecture is built on two encoder-decoder structures~(also termed as contracting-expanding structures)~\cite{noh2015learning,badrinarayanan2015segnet}, which are neural networks with an encoder~(contracting part) and a decoder~(expanding part). The encoder is designed to extract a feature representation of the input~(image) and the decoder translates the feature back to the input~(autoencoder~\cite{bourlard1988auto,hinton2006reducing}) or the desired output space. This structure is widely used in image restoration~\cite{bigdeli2017deep,mao2016image,ulyanov2017deep}, image transformation~\cite{hinton2011transforming,gonzalez2018image}, pose estimation~\cite{newell2016stacked} or semantic segmentation~\cite{noh2015learning,badrinarayanan2015segnet}. In addition, it has also been utilized for unsupervised feature learning~\cite{pathak2016context} and factor disentangling~\cite{mathieu2016disentangling,szabo2017challenges,hu2017disentangling}. In our framework, this architecture is adopted for both the image reconstruction module and the hand joint localization module, while we propose a novel unsupervised training method to ensure the separation of factors generated by the encoders.

\textbf{Learning Disentangled Representations}. Disentangling the factors of variation is a desirable property of learned representations~\cite{bengio2013representation}, which has been investigated for a long time~\cite{tenenbaum1997separating,tenenbaum2000separating,hinton2011transforming,hu2017disentangling}.
In~\cite{hinton2011transforming}, an autoencoder is trained to separate a translation invariant representation from a code that is used to recover the translation information. In \cite{rifai2012disentangling}, the learned disentangled representations is applied to the task of emotion recognition.
Mathieu~\etal combine a Variational
Autoencoder~(VAE) with a GAN to disentangle representations depending on what is specified (\ie
labeled in the dataset) and the remaining unspecified factors of variation~\cite{mathieu2016disentangling}.

Recently, factor disentanglement has also been used to improve visual quality of synthesized/reconstructed images and/or to improve recognition
accuracy for research problems such as pose-invariant face recognition~\cite{tran2017disentangled,peng2017reconstruction}, 
identity-preserving image editing~\cite{huang2017beyond,lample2017fader,li2016deep}, 
and hand/body pose estimation~\cite{ma2018disentangled,balakrishnan2018synthesizing}. 
However, these factor disentanglement methods usually either require paired data or explicit attribute supervision to encode the expected attribute. 
Two recent techniques, $\beta$-VAE~\cite{higgins2016beta} and DIP-VAE~\cite{kumar2017variational}, build on variational autoencoders (VAEs) to disentangle interpretable factors in an unsupervised way. However, they learn it by matching to an isotropic Gaussian prior, while our method learns disentanglement using a novel cycle-consistency loss.
\cite{baek2018augmented} improves the robustness of pose estimation methods by synthesizing more images from the augmented skeletons, which is achieved by obtaining more unseen skeletons instead of leveraging the unseen combinations of the specified factor~(pose) and unspecified factors~(background) in the existing dataset like ours. 
The most related work is \cite{yang2018disentangling}, which proposes an disentangled VAE to learn the specified~(pose) and additional~(appearance) factors. However, our method explicitly makes the appearance factor orthogonal to the pose during training process, while [2] only guarantees that the pose factor does not contain information about the image contents.


\section{Learning Self-Disentanglement}

In this section, we present our self-disentanglement framework. An overview of
the framework can be found in Fig.~\ref{fig:overview}.  Our framework encodes a
monochrome hand image into two orthogonal latent features: the pose feature
$\mathbf{p}$ and the appearance feature $\mathbf{a}$ using pose encoder~$E_p$ and appearance encoder~$E_I$. Without explicit
supervision on how these two features disentangle, we introduce the following
consistency criteria for self supervision.

\begin{figure}[t!]
	\centering
	\includegraphics[width=0.48\textwidth]{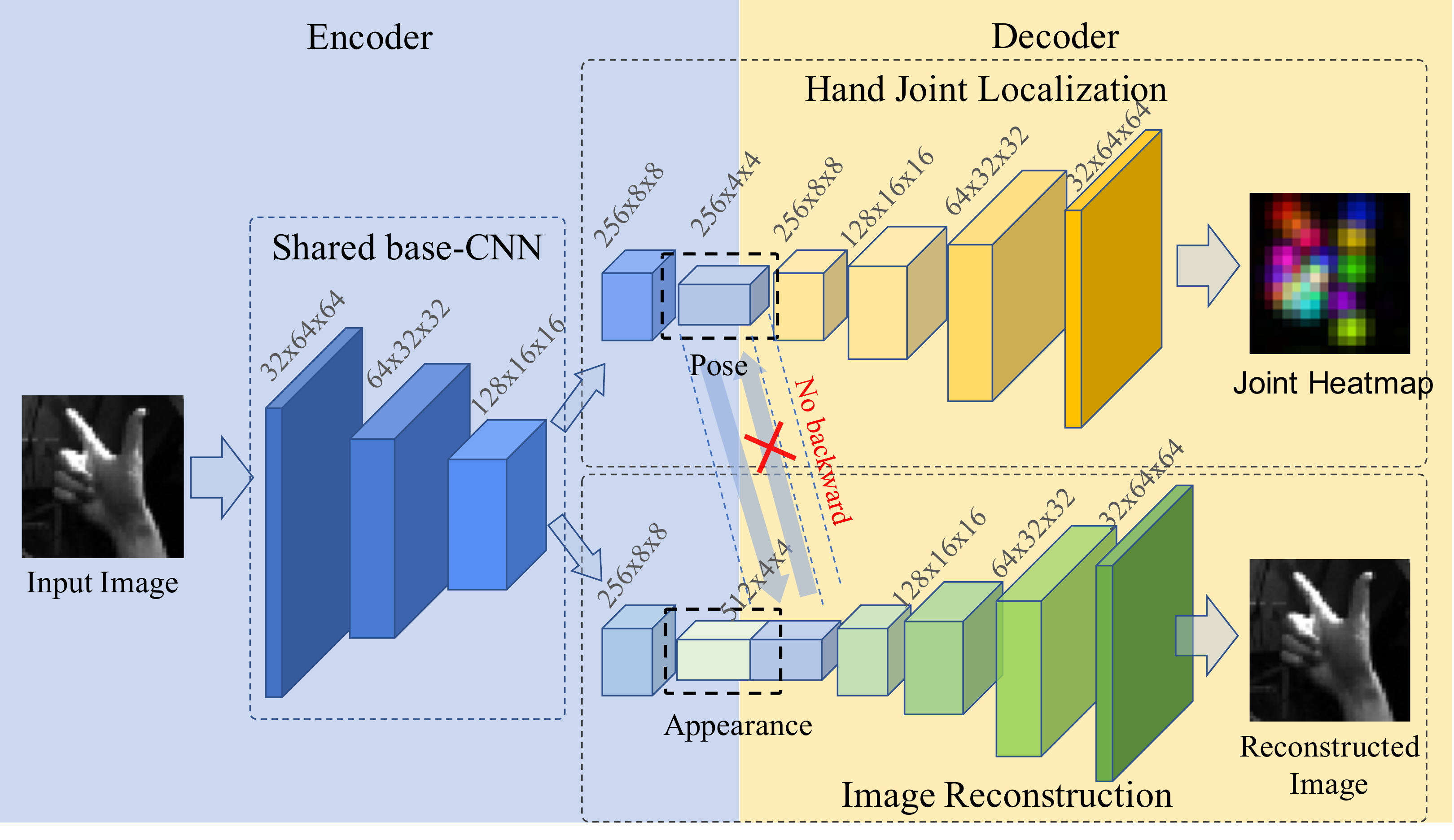}
	\caption{The structure of the hand joint localization module~(pose) and the image reconstruction module. 
		Both modules share some early-stage convolutional layers of the encoder. 
		Image decoder utilizes both the pose and appearance factors to reconstruct the image, 
		but the gradients back-propagated from the image reconstruction branch does not go backward to the pose factor learning.}	
	\label{fig:baseline}
	\vspace{-5mm}
\end{figure}

\subsection{Pose Estimation Loss}\label{sec:pose}

To encode the pose feature, we use a model similar to the contracting-expanding
structure of UNet~\cite{ronneberger2015u}. As shown in the top of
Fig.~\ref{fig:baseline}, we use down-sampling layers~(the pose encoder~$E_p$) to
encode the image $I^{(x)}$ into a latent pose feature $\mathbf{p}^{(x)}$. The
up-sampling layers~(the pose decoder~$D_p$) then decode $\mathbf{p}^{(x)}$ into a set of
hand joint heatmaps $\hat{P}^{(x)}$. Each heatmap $P^{(x)}$ is a Gaussian
centered at a single hand joint location~\cite{tompson2014real}. An L1 loss
penalizes differences between the predicted heatmaps $\hat{P}^{(x)}$ and the
ground truth heatmaps $P^{(x)}$:
\begin{equation}
L_{p}^{(x)} = \mathbb{E}\left[\norm{\hat{P}^{(x)} - P^{(x)}}_{1}\right]
\label{eq:loss_pose}
\end{equation}

Note that while skip connections are commonly used to preserve details in the output~\cite{ronneberger2015u}, we avoid these connections here, as they allow the network to bypass the latent pose feature, thus preventing proper disentanglement.  

\subsection{Image Reconstruction Loss}\label{sec:autoencoder}
To generate the appearance feature $\mathbf{a}^{(x)}$, we use another encoder-decoder network $E_{I}$ with the same contracting-expanding structure~(lower part of Fig.~\ref{fig:baseline}). This encoder shares the early-stage layers with the pose module $E_{p}$ as shown in Fig.~\ref{fig:baseline}.
To ensure that the two latent factors contain the information we expect, an
image reconstruction loss is introduced in the framework. The decoder network
$D_{I}$ now takes both the pose feature $\mathbf{p}^{(x)}$ and the appearance
feature $\mathbf{a}^{(x)}$ to reconstruct the original image as
${\hat{I}^{(x)}}$.  Supervision is provided by a reconstruction loss: we
penalize the difference between the decoded image $\hat{I}^{(x)}$ and the
original image $I^{(x)}$ using an L1 loss:
\begin{equation}
L_{I}^{(x)} = \mathbb{E}\left[\norm{\hat{I}^{(x)} - I^{(x)}}_1\right]
\label{eq:loss_reconstruct}
\end{equation}

In addition, a GAN loss~\cite{goodfellow2014generative} is used to encourage
the reconstructed image to be indistinguishable from the real hand images. The 
discriminator and generator losses are defined as follows:
\begin{equation}
\begin{split}
L_{D}^{(x)} & = \mathbb{E}\left[\log\left( D(I^{(x)}) \right)\right] + \mathbb{E}\left[\log\left(1-D(\hat{I}^{(x)})\right)\right] \\
L_{G}^{(x)} & = \mathbb{E}\left[\log\left(D(\hat{I}^{(x)})\right)\right]
\label{eq:loss_gan_d}
\end{split}
\end{equation}
where $L_{D}$ and $L_{G}$ denote the losses for discriminator and generator respectively.

One risk when using a reconstruction loss is that the network can ``hide'' appearance information in the encoded pose feature in order to improve the quality of the reconstruction.  This is contrary to our goal that the pose feature should solely encode an abstract representation of the pose alone.  To prevent this, during training, we block the gradients from the image reconstruction loss from back-propagating to the pose feature~(Fig. \ref{fig:baseline}); as a result, the pose encoder is not supervised by the image reconstruction loss, and thus has no incentive to encode appearance-related features.

\subsection{Learning Orthogonal Factors with Mix-Reconstruction}
Ideally, the extracted pose and appearance factor should be orthogonal to each other, that is, $\mathbf{a}$ and $\mathbf{p}$ should encode different aspects of the image.  This would allow combining any arbitrary pose/appearance pair to generate
a valid image.  
However, the autoencoder in Sec.~\ref{sec:autoencoder} has no incentive to keep the appearance factor separate from the pose
factor; the image reconstruction step works even if the appearance factor also encodes the pose.  

Previous work on factor disentanglement~\cite{gonzalez2018image,mathieu2016disentangling,peng2017reconstruction,liu2018exploring} uses image
pairs as supervision.  If we have two images that vary in appearance but have the same object category, then we could
use this to help the network learn what ``appearance'' means.  Nevertheless, in our case, we do not have such data pairs:
images that have identical pose but different lighting are difficult to obtain. Hence, factor disentanglement should be done without any knowledge of the data except the hand joint locations.

As shown in Fig.~\ref{fig:overview}, we appeal to a randomly sampled instance,
$I^{(y)}$, which has no relation to $I^{(x)}$ in either pose or
appearance~(different pose icons and background patterns denote the different
pose and appearance). We can extract the pose feature and appearance feature
$\mathbf{p}^{y}$ and $\mathbf{a}^{y}$ from the random instance $I^{(y)}$. Then
we concatenate $\mathbf{p}^{(x)}$ and $\mathbf{a}^{(y)}$, and use the decoder
$D_{I}$ to generate a novel ``mix-reconstructed'' image
$\hat{I}^{(xy)}$, which ideally combines the pose from $I^{(x)}$ and appearance
from $I^{(y)}$. 
$\hat{I}^{(xy)}$ is expected to have $I^{(x)}$'s pose and $I^{(y)}$'s
appearance, but there exists no image in our training set that embodies this
particular combination of pose and appearance. We cannot supervise the
reconstruction of $\hat{I}^{(xy)}$ directly. Consequently, we rely on cycle
consistency to provide indirect supervision.

\subsection{Cycle Consistency Loss}\label{sec:cycle}

To tackle the problem mentioned above, we further decode  $\hat{I}^{(xy)}$ back to $\hat{\mathbf{p}}^{(x)}$ and $\hat{\mathbf{a}}^{(y)}$ using the pose and appearance encoder as Sec.~\ref{sec:pose} and \ref{sec:autoencoder}. As shown in Fig.~\ref{fig:overview} (c), we re-combine the reconstructed factors $\hat{\mathbf{p}}^{(x)}$ and $\hat{\mathbf{a}}^{(y)}$ with $\mathbf{a}^{(x)}$ and $\mathbf{p}^{(y)}$ respectively to synthesize the original image as $\tilde{I}^{(x)}$ and $\tilde{I}^{(y)}$. Now we build a disentangle-mix-disentangle-reconstruct cycle to generate back to the original input~(denoted as the self-disentanglement), and we use the following cycle consistency losses during training:

\begin{equation}
\begin{split}
L_{cycle-img}^{(x)} & = \mathbb{E}\left[\norm{\tilde{I}^{(x)} - I^{(x)}}_1\right] \\
L_{cycle-img}^{(y)} & = \mathbb{E}\left[\norm{\tilde{I}^{(y)} - I^{(y)}}_1\right] \\
\label{eq:loss_cycle}
\end{split}
\end{equation}
\vspace{-5mm}

The reconstructed pose factors $\hat{\mathbf{p}}^{(x)}$ and $\hat{\mathbf{a}}^{(y)}$ should also match the $\mathbf{p}^{(x)}$ and $\mathbf{a}^{(y)}$. An additional dual feature loss is also added as an auxiliary supervision to enforce the feature-level consistency:
\begin{equation}
\begin{split}
L_{dual-pose}^{(x)} & = \mathbb{E}\left[\norm{\hat{\mathbf{p}}^{(x)} - \mathbf{p}^{(x)}}_1\right] \\
L_{dual-img}^{(y)} & = \mathbb{E}\left[\norm{\hat{\mathbf{a}}^{(y)} - \mathbf{a}^{(y)}}_1\right] \\
\label{eq:loss_dual}
\end{split}
\end{equation}
\vspace{-1cm}

where $\mathbf{p}^{(x)}$ and $\mathbf{a}^{(y)}$ here only serve as fixed training targets, and the gradients are not back-propagated through to $E_{p}$ and $E_{I}$.

In addition, the mix-reconstructed image $\hat{I}^{(xy)}$ is also expected to output the pose from $I^{x}$. Therefore, as shown in Fig.~\ref{fig:overview}~(d) the reconstructed pose code $\hat{\mathbf{p}}^{(x)}$ is decoded with the pose decoder $D_{p}$ to the hand joint heatmap $\tilde{P}^{(x)}$, which should match the original heatmap $P^{(x)}$:
\begin{equation}
L_{cycle-pose}^{(x)} = \mathbb{E}\left[\norm{\tilde{P}^{(x)} - P^{(x)}}_1\right]
\label{eq:loss_cycle_pose}
\end{equation}

\subsection{End-to-end Training}\label{sec:end-to-end}
The model is trained end-to-end with randomly sampled pairs $I^{(x)}$ and $I^{(y)}$:
\begin{equation}
\begin{split}
Loss = & L_{p} + \alpha_1 L_{I} + \alpha_2 L_{D} + \alpha_3 L_{G} \\
&  + \alpha_4 L_{cycle-img} + \alpha_7 L_{cycle-pose} \\
& +\alpha_6 L_{dual-img} + \alpha_5 L_{dual-pose} \\
\label{eq:loss_total}
\vspace{-10pt}
\end{split}
\end{equation}
where $L_{(\cdot)}$ denotes the sum of the corresponding losses for the pairs $L_{(\cdot)}^{(x)}$ and $L_{(\cdot)}^{(y)}$.

When evaluating cycle consistency, the pose decoder $D_p$ and the image decoder
$D_I$ serve as an evaluator to estimate whether the mix-reconstructed image
$\hat{I}^{(xy)}$ can output the correct
hand joint heatmap and can be encoded into expected features. We don't necessarily want to train them based on the mix-reconstructed image because it
may be poor in quality, especially during the early stage of training.  
Therefore we fix the parameters in these two decoders in
Fig.~\ref{fig:overview}~(c-d) (shown with dash outline). They are simply a copy
of the modules in Sec.~\ref{sec:pose} and Sec.~\ref{sec:autoencoder}, but do
not accumulate gradients in back-propagation. This simple strategy greatly 
stablilizes training.

\begin{table}[h]
	\renewcommand{\arraystretch}{1.2}
	\setlength{\tabcolsep}{5pt}
	\begin{center}
		\begin{tabularx}{0.85\linewidth}{l | c | c}
			\hline
			Dataset & Train (\#frames)& Testing (\#frames)\\
			\hline
			Real & 123,034  & 13,416 \\
			
			Synthetic & 123,034 $\times$ 2  & 13,416 $\times$ 2\\
			\hline
		\end{tabularx}
	\end{center}
	\caption{Statistics of the real and synthetic hand image datasets. 
		The synthetic dataset is made up of pairs of images, 
		which share the pose but differ in backgrounds, lighting conditions and the hand textures.}
	\label{tab:dataset}
\end{table}

\section{Experiments}

\subsection{Data Preparation}
We collect a dataset of monochrome hand
images captured by headset-mounted monochrome cameras in a variety of environments and lighting conditions. To obtain
high quality ground truth labels of 3D joint locations, we rigidly attach the
monochrome camera to a depth camera, then apply~\cite{tompson2014real} on the
depth image to obtain 3D key points on the hand. With careful camera
calibration, we transform the 3D key points to the monochrome camera space as
ground truth labels. The training images are then generated as a 64x64 crop
around the hand. 

In addition, we render a synthetic dataset of hand images using the same hand
poses from the monochrome dataset. Each pose is rendered into a pair of images
with different environment maps, lighting parameters and hand textures. This synthetic dataset
offers perfectly paired data of the same pose with different appearances.
Tab~\ref{tab:dataset} shows statistics of the two datasets.

\begin{figure*}[t!]
	\centering
	\subfigure[Real Images]{\label{fig:a}\includegraphics[width=0.45\textwidth]{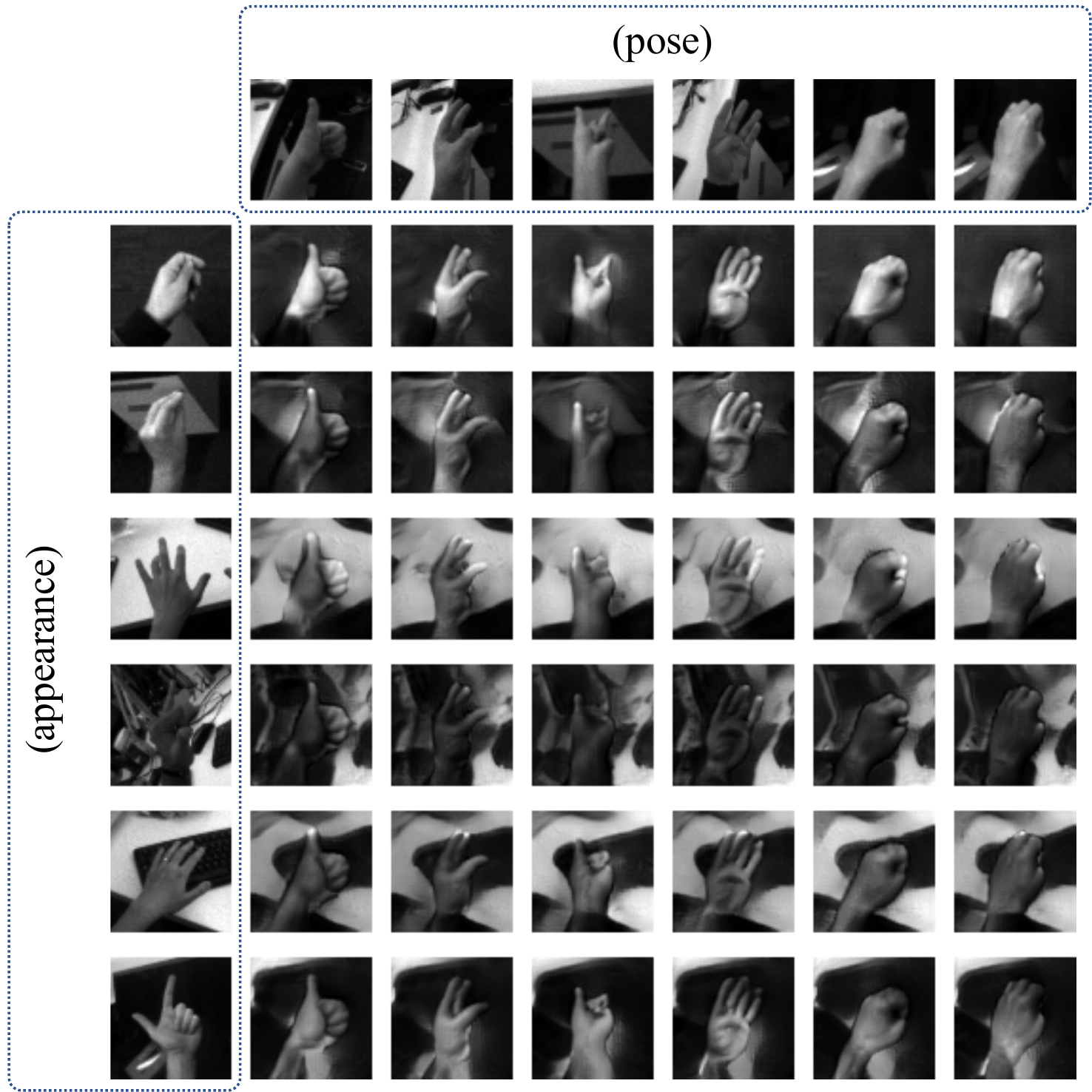}}
	\subfigure[Synthetic Images]{\label{fig:b}\includegraphics[width=0.45\textwidth]{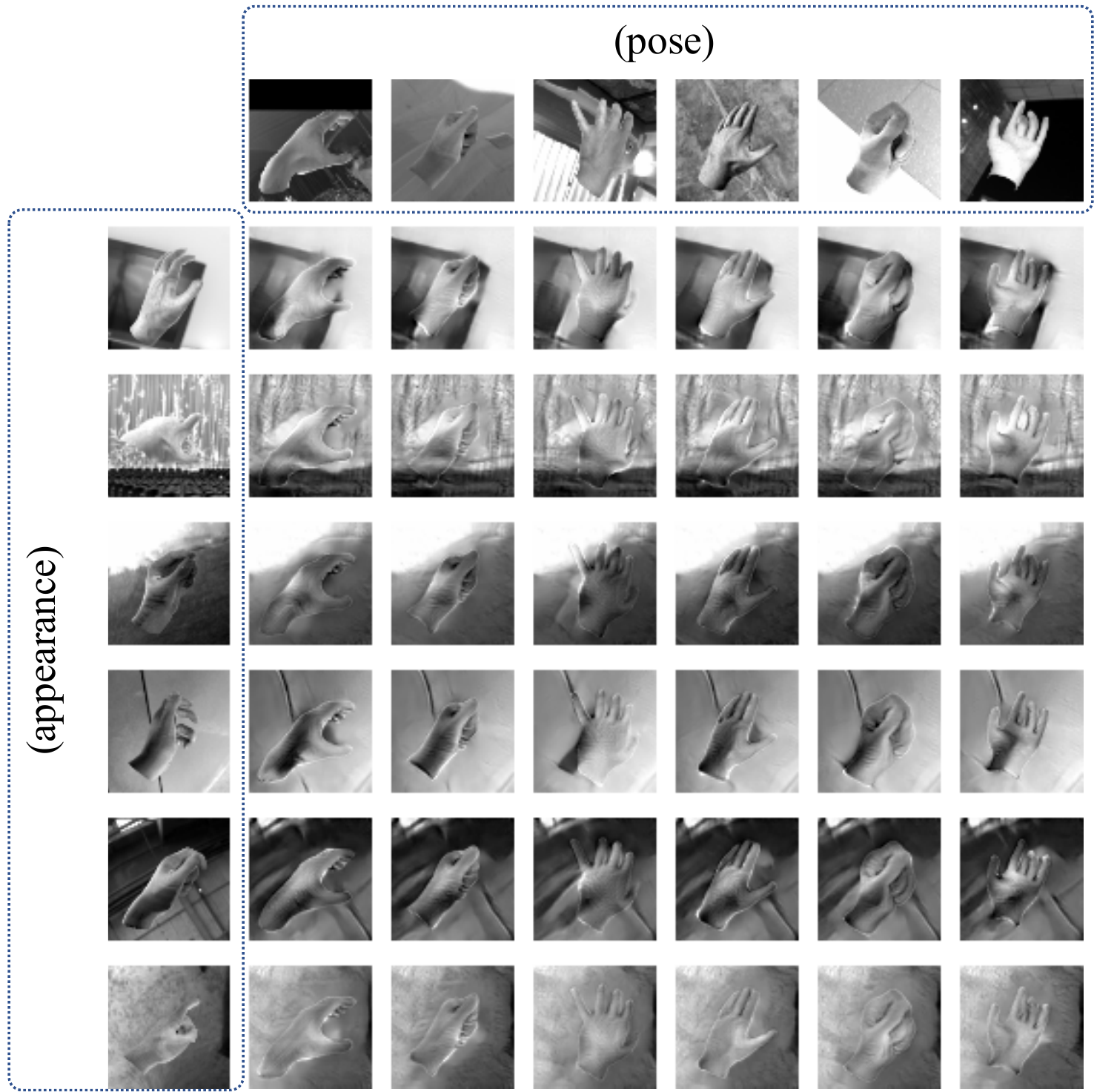}}
	\caption{Self-disentanglement on Real~(left) and Synthetic~(right) data. The image on the top row provide the ``pose'' while the image on the left-most column offers the ``appearance''. The images in the middle matrix are generated with our proposed method using the corresponding pose and the row-wise appearance.}	
	\label{fig:qualitative}
	\vspace{-10pt}
\end{figure*}

\subsection{Implementation Details}
We use an encoder-decoder architechure following UNet~\cite{ronneberger2015u}
without skip-connections as the base model. The encoder is shared between the
pose feature $E_{I}$ and the appearance feature $E_{p}$ before the last
downsampling layer. Two different decoders are used in the Hand Joint
Localization branch and the Image Reconstruction branch respectively, where the
image reconstruction branch decodes from both the pose feature and the
appearance feature (Fig~\ref{fig:baseline}). 

Both the encoder and the two decoders have a depth of 4. Each block in the
encoder consists of repeated application of two 3x3 convolutions, each followed
by a rectified linear unit (ReLU) and a 2x2 max pooling operation with stride 2
for down-sampling. The pose decoder $D_{p}$ employs a 2x2 deconvolution layer
in each block for up-sampling, while the image decoder $D_{I}$ uses
nearest-neighbor upsampling followed by a 3x3 convolution instead to avoid the
checkerboard artifact~\cite{dong2016image}.  Fig.~\ref{fig:baseline}
illustrates the detailed model structure. At training time, we initialize all
parameters randomly, and use the Adam~\cite{adam} optimizer with a fixed
learning rate 0.001. A total of 75 epochs is run with a batch size of 128. 

\subsection{Orthogonal Feature Space From Self-Disentanglement}
\label{sec:qualitative}
We visually validate the orthogonality of the two feature spaces by
reconstructing novel images using the pose feature from one image and the
appearance feature from another. Fig.~\ref{fig:qualitative} shows a matrix of
generated results on both the captured dataset and the synthetic dataset. We
can successfully reconstruct the desired hand pose under different lighting and
background. For instance, the hands in the first two rows of
Fig.~\ref{fig:qualitative} (a) are lit by light source from the left, consistent in
appearance with the source images. Even though the network cannot reproduce all
the details in the background, it generates similar statistics. We refer readers 
to the supplementary video for more results.

There are still noticeable artifacts in the generated images, especially when
the pose estimator does a poor job either in the appearance image (row 2 in
Fig.~\ref{fig:qualitative}(b)) or in the pose image (column 3 in
Fig.~\ref{fig:qualitative}(a)). Interestingly, because we don't have any key points
on the arm, it is encoded into the appearance feature by our network (row 6
in Fig.~\ref{fig:qualitative}(a)). 

\begin{table*}[h!]
	\setlength{\tabcolsep}{0.1pt}
	\tiny
	\begin{center}
		\begin{tabular}{| c | l  |}
			\hline
			\tabincell{c}{Appearance\\ Factor} &
			\raisebox{-0.40\totalheight}{\includegraphics[width=0.85\textwidth]{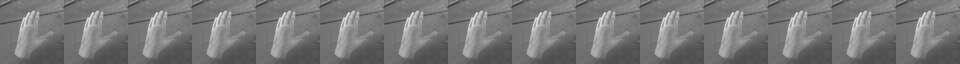}} \\\hline
			\tabincell{c}{Pose\\ Factor} & 
			\raisebox{-0.35\totalheight}{\includegraphics[width=0.85\textwidth]{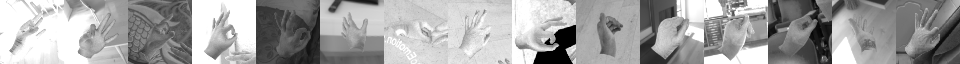}} \\\hline
			AutoEncoder~\cite{masci2011stacked} & 
			\raisebox{-0.35\totalheight}{\includegraphics[width=0.85\textwidth]{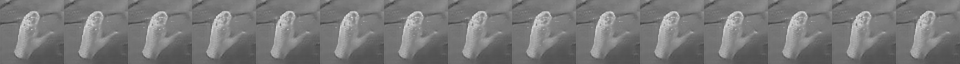}} \\\hline
			Paired Data~\cite{gonzalez2018image} &
			\raisebox{-0.35\totalheight}{\includegraphics[width=0.85\textwidth]{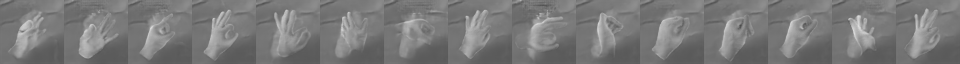}} \\\hline
			Ours & 
			\raisebox{-0.35\totalheight}{\includegraphics[width=0.85\textwidth]{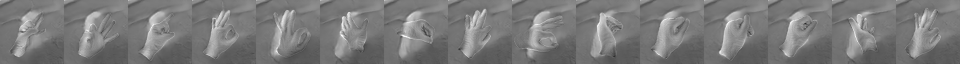}} \\\hline
			\hline
			\tabincell{c}{Appearance\\ Factor} &
			\raisebox{-0.35\totalheight}{\includegraphics[width=0.85\textwidth]{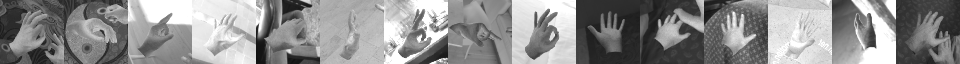}} \\\hline
			\tabincell{c}{Pose\\ Factor} & 
			\raisebox{-0.35\totalheight}{\includegraphics[width=0.85\textwidth]{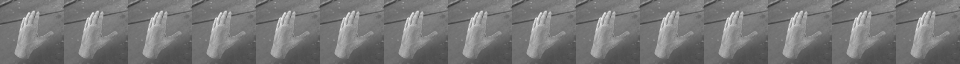}} \\\hline
			AutoEncoder~\cite{masci2011stacked} & 
			\raisebox{-0.35\totalheight}{\includegraphics[width=0.85\textwidth]{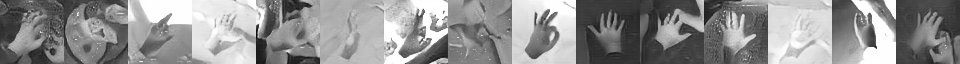}} \\\hline
			Paired Data~\cite{gonzalez2018image} & 
			\raisebox{-0.35\totalheight}{\includegraphics[width=0.85\textwidth]{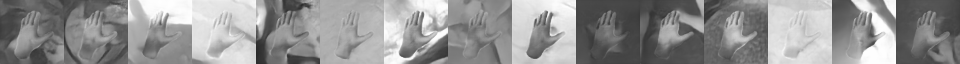}} \\\hline
			Ours & 
			\raisebox{-0.35\totalheight}{\includegraphics[width=0.85\textwidth]{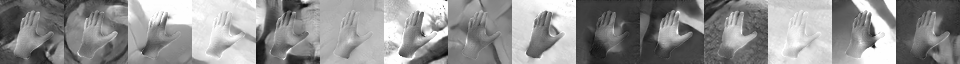}} \\\hline
		\end{tabular}
	\end{center}
	\caption{Comparison with the existing methods on the paired synthetic data. Top part: fixed appearance factors with varying the pose ones. Bottom Part: varying pose factors with fixed appearance ones.  \textbf{Appearance Factor} shows the images providing the appearance factors. \textbf{Pose Factor} shows the images providing the appearance factors. \textbf{AutoEncoder} denotes the image reconstruction along with the pose module shown in Sec.~\ref{sec:autoencoder}. \textbf{Paired Data} denotes the factor disentanglement using paired data~\cite{gonzalez2018image}. \textbf{Ours} is our proposed self-disentanglement without leveraging the paired data.}
	\label{tab:comparison}
	\vspace{-10pt}
\end{table*}

\begin{figure}[t!]
	\centering
	\includegraphics[width=0.45\textwidth]{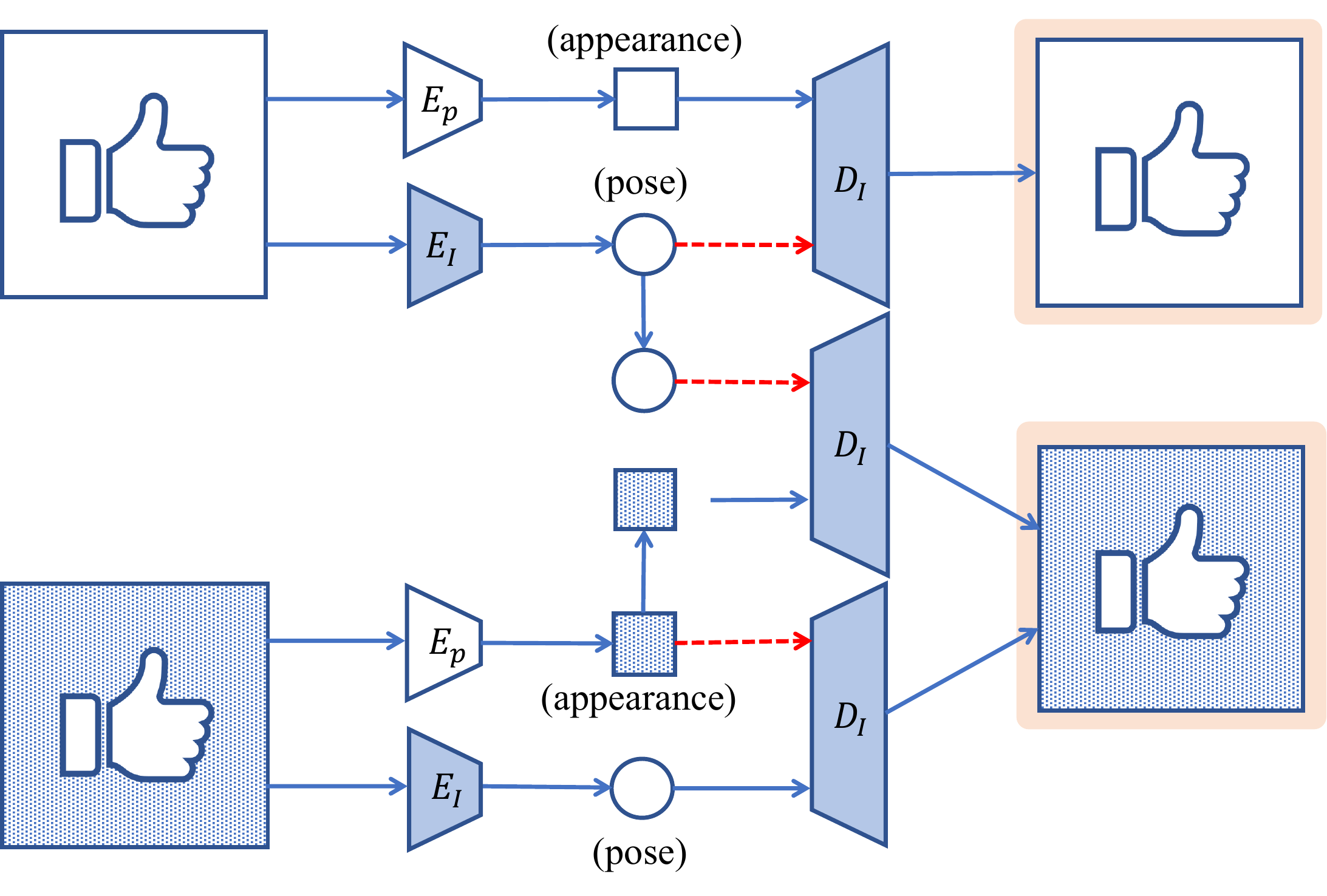}
	\caption{Factor Disentanglement with paired data~\cite{gonzalez2018image}. The two inputs share the pose but differ in image appearance.}	
	\label{fig:model_baseline}
	\vspace{-12pt}
\end{figure}

\begin{table}[h]
	\renewcommand{\arraystretch}{1.2}
	\setlength{\tabcolsep}{5.2pt}
	\begin{center}
		\begin{tabularx}{\linewidth}{l | c | c }
			\hline
			Dataset & Paired Data~\cite{gonzalez2018image} & Ours\\
			\hline
			I.S.~\cite{salimans2016improved}  &  4.96 $\pm$ 0.11 & 5.10 $\pm$ 0.10\\
			\hline 
			Preference & 51.66\% (529 / 1024) & 48.33\% (495 / 1024) \\
			\hline
		\end{tabularx}
	\end{center}
	\caption{Quantative Comparison of the factor disentanglement using Paired Data~\cite{gonzalez2018image} and our proposed self-disentanglement, including Inception Scores~\cite{salimans2016improved}~(\textbf{I.S.}) and User Study.}
	\label{tab:inception_score}
	\vspace{-12pt}
\end{table}

\subsection{Comparison to Supervised Disentanglement}
To prove the effectiveness of our proposed self-disentanglement, 
we compare our method with two baselines: 
(1) the Auto-Encoder~\cite{masci2011stacked} with the structure shown in Sec.~\ref{sec:autoencoder}; 
(2) the factor disentanglement~\cite{gonzalez2018image} using the paired data that have identical pose but different appearance. 
Detailed experimental results are shown in Tab.~\ref{tab:comparison}. 

We can see that the images from the \emph{Appearance Factor} row and the Autoencoder results row in Tab.~\ref{tab:comparison}, are nearly the same. It shows that, without supervision on orthogonality, the Auto-Encoder model encodes the entire input
image into the appearance feature, and discards the pose feature in decoding. Therefore, the pose and appearance factors are not fully disentangled. 
Checking the results of disentanglement with paired data~\cite{gonzalez2018image} and our self-disentanglement, both methods are able to
combine pose feature and appearance feature from two different source images to
construct a novel image with specified pose and appearance. 
Our model generates visually similar images to the model trained with paired data.

Furthermore, we randomly swap the hand and appearance factors of the held-out set to generate a new set of images, and then calculate the inception scores~\cite{salimans2016improved} and perform a user study on the preference of between our method and \cite{gonzalez2018image} in Tab.~\ref{tab:inception_score}. The comparable results validate our claims. 

\begin{table*}[h]
	\renewcommand{\arraystretch}{1.2}
	\setlength{\tabcolsep}{8pt}
	\begin{center}
		\begin{tabularx}{0.95\linewidth}{c| l | c | c | c}
			\hline
			ID & Model & Epochs  & MSE (in pixels) & Improvements\\
			\hline
			1 & Baseline Pose Estimator 							&	75	& 4.174  & - \\
			2 & Pose Estimator + Image Reconstruction 				&	75	& 3.982  & 4.60\% \\
			3 & Our proposed Self-Disentanglement 				&	75	& 3.923  & 6.02\% \\
			4 & Our proposed Self-Disentanglement + Resume (**) &	150	& 3.864  & 7.44\% \\
			5 & (**) + No Pose Estimator Detach 					&	150	& 3.756  & 10.02\%\\
			6 & (**) + No Pose Estimator Detach + no Pose Detach 	&	150	& \textbf{3.735}  & \textbf{10.53\%} \\
			\hline
		\end{tabularx}
	\end{center}
	\caption{Ablation Study of the influence brought by Self-Disentanglement training on Hand Joints Localization. Mean Square Error~(MSE) between the predicted location and the ground-truth is used to evaluate the accuracy, which is the lower the better. All models use the same model structure. \textbf{Resume} denotes resuming the training for another 75 epochs. \textbf{No Pose Estimator Detach} means when resuming training, the pose estimator will get trained on the mix-reconstructed images. \textbf{No Pose Detach} means when resuming training, the loss back-propagated from the image generation branch will go backward to the pose estimator through the pose factor.  }
	\label{tab:quantitative}
	\vspace{-12pt}
\end{table*}

\subsection{Improve Pose Estimation with Disentanglement}
An important application of our disentanglement model is to improve the
robustness of the pose estimation module. We examine how each criterion in the
disentanglement process affect the pose estimator step by step.

We fit the predicted heatmap of every joint to a gaussian distribution, and 
use the mean value as the predicted locations of the joints. 
Tab.~\ref{tab:quantitative} shows quantitative results, where the MSE denotes 
the mean square error of the predictions in pixels. The baseline pose estimator
is trained with supervised learning (Sec~\ref{sec:pose}). When we add the image
reconstruction loss (Sec~\ref{sec:autoencoder}, the accuracy is already
improved by 4.60\%. It suggests that the image reconstruction task encourages the
shared base layers (Fig.~\ref{fig:baseline}) to extract more meaningful low
level features for pose estimation. Adding the cycle consistency loss
(Sec~\ref{sec:cycle}) further boosts the performance by 6.02\%.

In Sec~\ref{sec:end-to-end}, we employe a strategy to stabilize training by
disabling back-propagation to the pose feature (\emph{Pose Detach}) as well as
the back-propagation to the pose estimator parameters (\emph{Pose Estimator
Detach}). This is useful because the most reliable supervision is from the
joint location labels, and we don't want to distract the pose estimator by
auxiliary tasks that are more ambiguous in the early stage.  However, once we
have a reasonable disentanglement network, the additional supervision from
image reconstruction and cycle consistency may help the pose estimator to
better differentiate a pose from its appearance. We conduct two additional
experiments using warm start from Model 3 to test this hypothesis. The new
baseline trains our network as described in Sec~\ref{sec:end-to-end} for
another 75 epochs (Model 4). The first experiment allows back-propagation to
the pose feature (Model 5), and the second experiment allows back-propagation
to both the pose feature and the pose estimator parameters (Model 6). Both
models are trained from Model 3 for 75 epochs.  While the pose estimator
benefits from warm start and the additional epochs, we can observe even greater
improvements in accuracy when back-propagation is enabled. These two
experiments demonstrate the effectiveness of self-disentanglement in improving
the robustness of pose estimation to make it more resilient to environment
variations.

\begin{figure}[t!]
\vspace{-8pt}
	\centering
	\subfigure[Retrieve with Pose]{\label{fig:a}\includegraphics[width=0.23\textwidth]{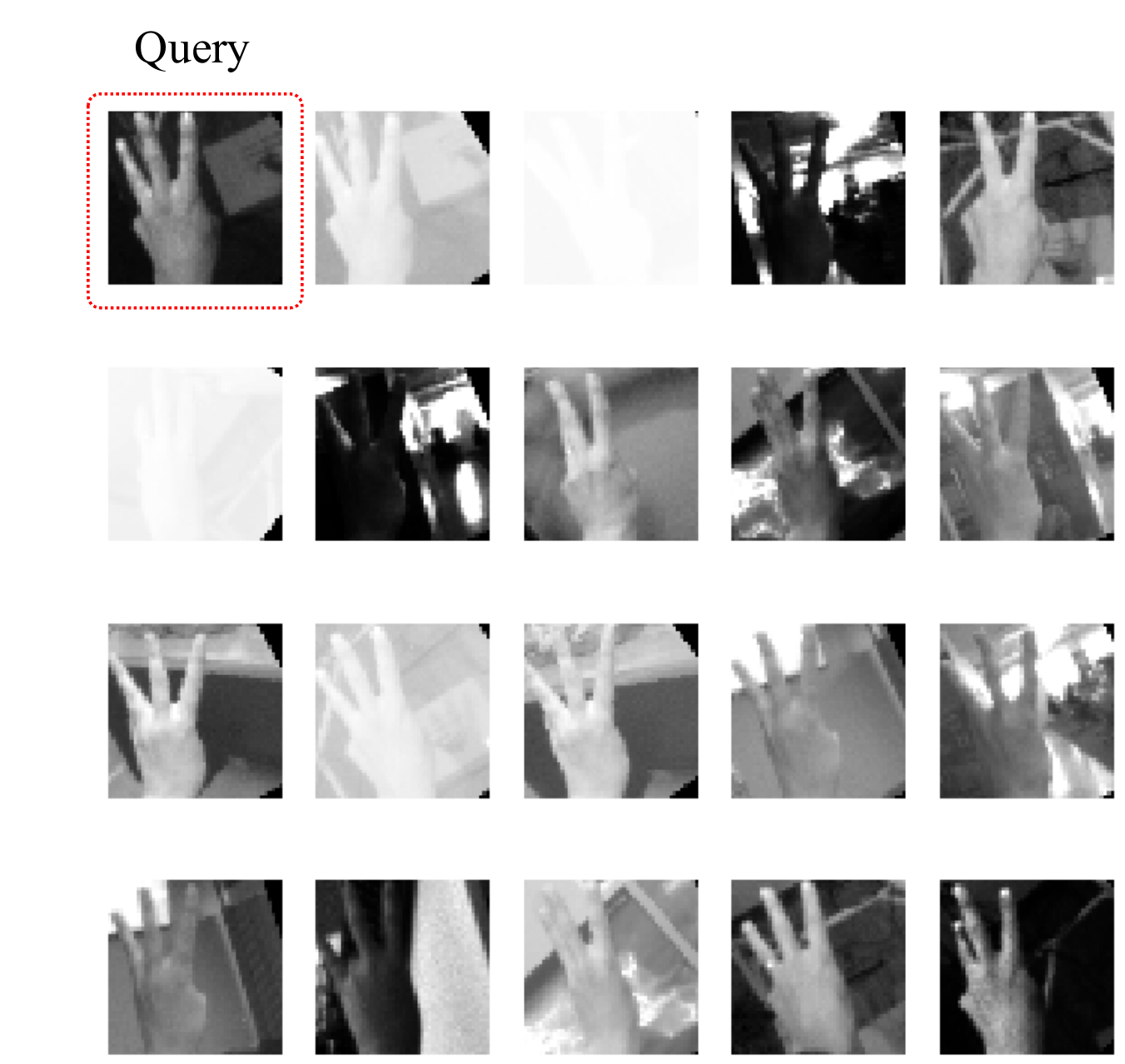}}
	\subfigure[Retrieve with Appearance]{\label{fig:b}\includegraphics[width=0.23\textwidth]{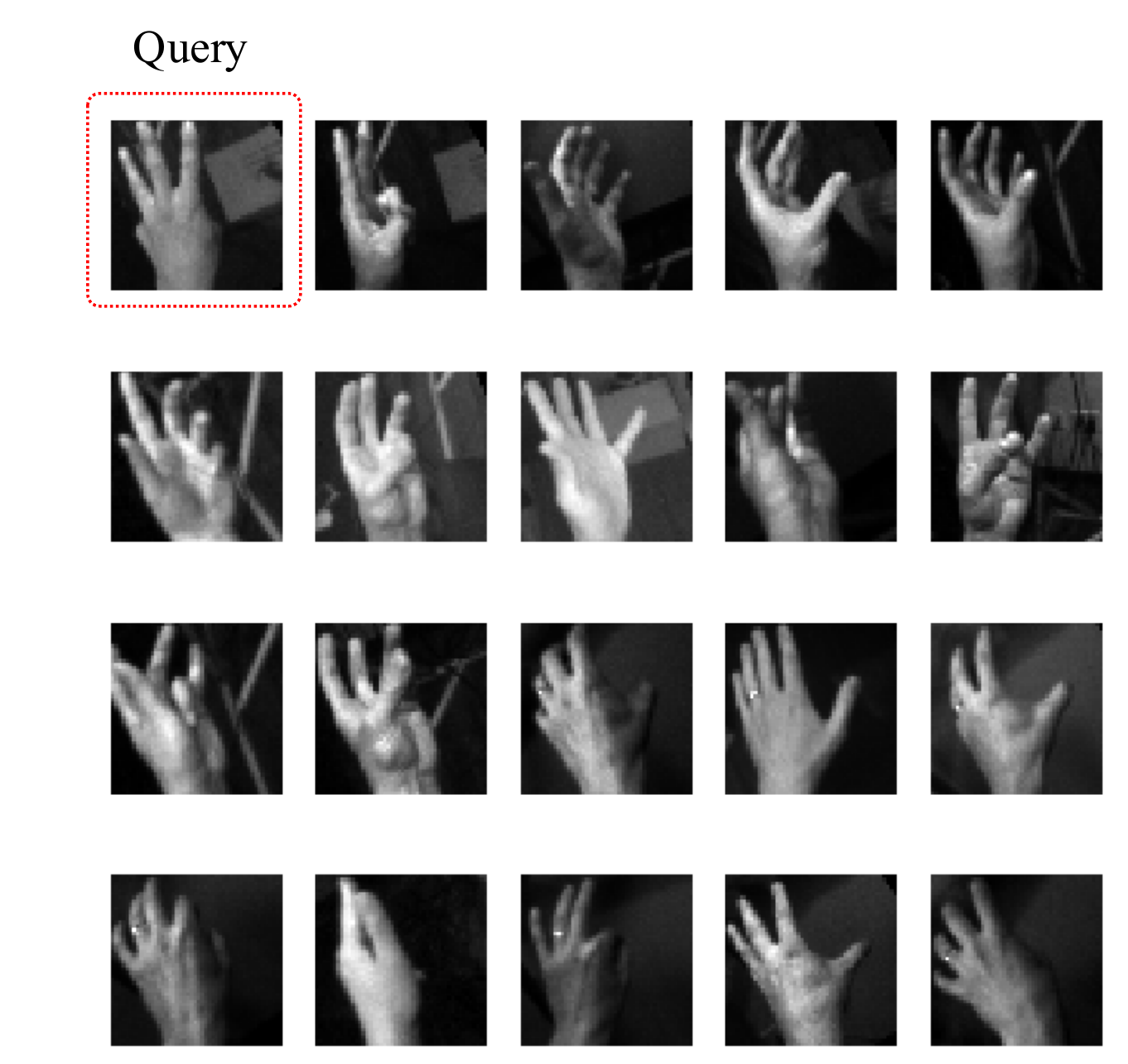}}
	\caption{Image retrieval with disentangled factors.}	
	\label{fig:retrieve}
	\vspace{-12pt}
\end{figure}

\subsection{Image Retrieval using Disentangled Factors} We can examine the
feature spaces by looking at images with similar features.  For instance, if we
query images with similar pose features, we will get images of similar hand
poses under different environment variations. Likewise, if we query images with
similar appearance features, we will get images in a similar environment but
with different hand poses. Fig.~\ref{fig:retrieve} shows the top-20 nearest
images from the monochrome dataset of the same query image in the pose space
and the appearance space respectively. The query results further confirm the
success of our method to disentangle the two factors.

\section{Discussion}
While we believe our method successfully disentangles pose features, we can
only indirectly validate the result by reconstructing novel images from random
pose-appearance feature pairs using a GAN framework. The reconstruction
captures the desired hand pose with consistent shading and background with the
environment, but not without artifacts. The most severe issues are usually
around the wrist and arm region, where the pose key points are sparse. Since
key points are the only direct supervision, the model needs to differentiate hand
pixels from background pixels based on the key points, and will make mistake where
the connection is weak. Incorporation of pixel label masks or dense annotations
as supervision to pose estimation and image reconstruction can potentially
improve the image quality. Another interesting failure case is when the pose
estimation makes a mistake, and the reconstruction image shows the wrongly
estimated pose rather than the original input pose. It shows that while we are
successful in disentanglement, there are other factors contributing to the
robustness of pose estimation. In the future, we would like to investigate a
more direct and quantitative measure of the effectiveness of disentanglement,
and to improve the quality of image reconstruction to enrich any
existing training dataset with a wide range of appearance variations.

\section{Conclusion}
In this paper, we present a self-disentanglement method to decompose a
monochrome hand image into representative features of the hand pose and its complementary features of the image appearance. Without the
supervision of paired images, we show that our method with cycle consistency principle is
sufficient to ensure orthogonality of the pose feature and the appearance
feature. Such flexibility makes our method applicable to any existing deep
learning based pose estimation framework without requiring additional data or
labels. When tested with a captured dataset of monochrome images, we
demonstrate significant improvement in the robustness of the pose estimator to
environment variations, comparing to a conventional supervised pose estimation
baseline. Additionally, compared to a disentanglement model learned from paired training
data, our model also performs similarly in terms of synthesized image quality proofing the success of
self-disentanglement.

{\small
\bibliographystyle{ieee}
\bibliography{egbib}
}

\end{document}